\documentclass[11pt]{article}

\usepackage[final]{acl}

\usepackage{times}
\usepackage{latexsym}

\usepackage[T1]{fontenc}

\usepackage[utf8]{inputenc}

\usepackage{microtype}

\usepackage{inconsolata}

\usepackage{makecell}
\usepackage{graphicx}
\usepackage{array}
\usepackage{booktabs}
\usepackage{tikz}
\usetikzlibrary{arrows.meta,calc,fit,positioning}
\definecolor{handoffgreen}{RGB}{0,110,45}
\definecolor{handoffred}{RGB}{170,45,45}
\definecolor{handoffgray}{RGB}{100,100,100}

\title{Handoff Debt: The Rediscovery Cost When Coding Agents Take Over Interrupted Tasks}

\author{
 \textbf{Dipesh KC\textsuperscript{1}},
 \textbf{Anjila Budathoki\textsuperscript{2}}
\\
\\
 \textsuperscript{1} Independent Researcher,
 \textsuperscript{2} University of Tennessee, Knoxville
\\
 \small{
   \href{mailto:kcdipesh429@gmail.com}{\texttt{kcdipesh429@gmail.com}},
   \href{mailto:abudatho@vols.utk.edu}{\texttt{abudatho@vols.utk.edu}}
 }
}

\begin{document}
\raggedbottom
\maketitle

\begin{abstract}
Coding-agent benchmarks evaluate whether a single uninterrupted agent can resolve a repository issue. Real software work is messier: tasks are interrupted, reassigned, reviewed, and resumed from partial states left by another agent or engineer. We study this missing dimension through \emph{handoff debt}: the rediscovery cost imposed when a predecessor's work is opaque or incomplete. Our takeover protocol interrupts a coding agent at deterministic handoff points, freezes the repository, and evaluates successor agents under four handoff views: repository state only, raw trace, summary notes, and structured notes. Across 75 source tasks, the protocol generates 181 handoff-point tasks and 724 takeover runs per successor model. Across three successor models, context-bearing handoffs reduce median agent events by 20--59\% and cumulative prompt tokens by 42--63\% relative to repository-only takeover. Solved-rate effects are smaller and model-dependent, but efficiency gains are consistent. These findings suggest that coding-agent evaluation should report not only whether a task is solved, but also how costly that work is for another agent to resume.\footnote{Code:\url{https://github.com/anjilab/agent-handoff-debt}}
\end{abstract}

\section{Introduction}

Recent software-engineering benchmarks have made coding agents measurable by asking whether, given a repository issue, an agent can produce a patch that passes the official tests. This abstraction is effective and reproducible for evaluating agents on real repositories \citep{jimenez2024swebench,yang2024sweagent}. However, this abstraction leaves out a common real-world case: \textit{takeover}, where one agent inherits an interrupted repository from another and must reconstruct what was changed, what was already attempted, and which intermediate artifacts can be trusted. In such handoffs, partial work is only valuable if the successor can understand it well enough to resume from it.


We call this cost \emph{handoff debt} and introduce a protocol to measure it. Handoff debt arises when an agent makes visible progress but leaves state that a successor cannot readily continue from, such as unexplained edits, scratch files, hidden assumptions, or missing validation evidence. A metric based solely on final resolution cannot distinguish between costly rediscovery and efficient continuation. Two predecessor agents may leave the same checkpointed repository, yet their successors can face very different continuation costs: one may continue immediately, while another must spend many tool interactions rediscovering intent from scratch files and incomplete command history.

We run experiments on SWE-bench Verified \citep{jimenez2024swebench,openai2024swebenchverified} using an OpenHands-style coding-agent environment \citep{openhands2024}. A predecessor agent begins a source task. We interrupt the agent at observable handoff points: \emph{After first source edit}, \emph{After first validation result}, or \emph{After first post-failure edit}. Each handoff point becomes a takeover task in which a successor resumes from the same checkpointed repository under four handoff views, meaning four ways of exposing predecessor context.

The four views differ in how much predecessor context the successor receives. \emph{Repository only} asks whether the filesystem state alone is enough. \emph{Raw trace} exposes the predecessor event log, but it is large, unstructured, and unbounded. \emph{Summary notes} test whether free-form compression can preserve intent and evidence from the trace. \emph{Structured notes} test whether a fixed continuation contract can make the handoff bounded and auditable. The central question is:

\emph{Which handoff view lets a successor agent correctly and efficiently resume interrupted coding work?}

Figure~\ref{fig:architecture} summarizes the resulting evaluation architecture. Our contributions are:

\begin{figure*}[t]
\centering
\includegraphics[width=\textwidth]{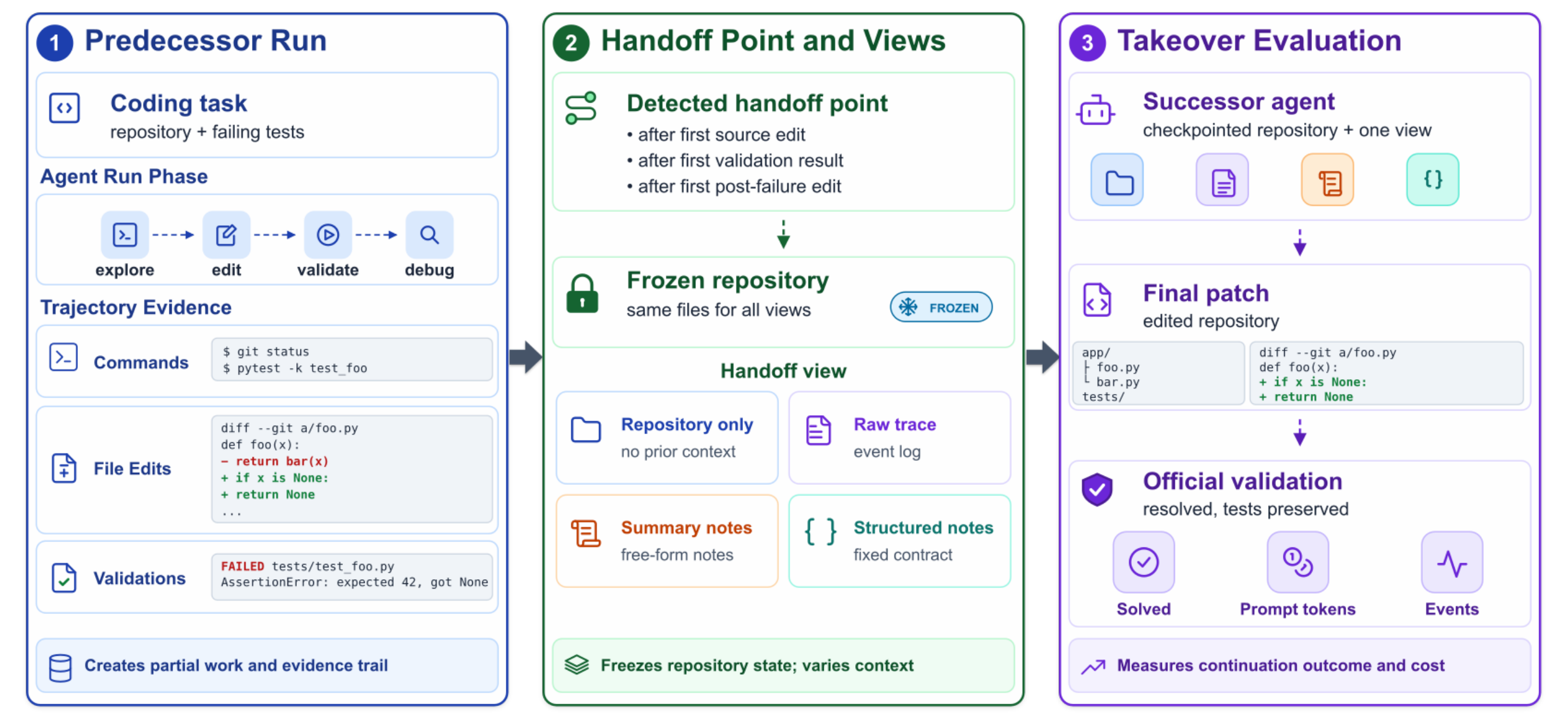}
\caption{\textbf{Handoff debt evaluation architecture.} A predecessor produces repository state and trajectory evidence before interruption. At a detected handoff point, the checkpointed repository is held fixed while the successor receives one of four handoff views. Final states are scored by official SWE-bench validation and efficiency metrics, including agent events and prompt tokens.}
\label{fig:architecture}
\end{figure*}

\begin{itemize}
    \item We formulate \emph{handoff debt} as a measurable property of coding-agent takeovers, capturing the cost of inheriting partial work.
    \item We introduce a takeover protocol that converts SWE-bench Verified tasks into handoff-point tasks, each evaluated under multiple handoff views.
    \item We compare four handoff views while holding the checkpointed repository fixed, isolating the effect of the handoff view.
    \item We propose a structured handoff schema as a bounded continuation contract, with fixed fields for changed files, validation evidence, uncertainty, rollback risk, and verification.
    \item We show that predecessor context primarily reduces rediscovery effort, cutting median agent events and prompt tokens even when solved-rate gains are modest.
\end{itemize}

\section{Problem Setup}

\subsection{Handoff Debt}

We use \emph{predecessor} for the agent that begins a task and \emph{successor} for the agent that resumes it after interruption. A \emph{handoff} is the transfer of partial-work state between them, and the successor's continuation episode is a \emph{takeover}. Let a predecessor agent $A$ operate on task $x$ and reach an intermediate checkpoint $t$. The checkpoint contains a repository state $s_t$ and a predecessor trajectory $\tau_{\leq t}$, including commands, file edits, observations, validation results, and model messages observed before the interruption. We later measure successor effort in \emph{agent events}, OpenHands trajectory records comprising LLM actions and tool observations. A successor agent $B$ receives the original task prompt, the checkpointed repository $s_t$, and a handoff view $h_t$ derived from some subset or transformation of $\tau_{\leq t}$.

The takeover succeeds if $B$ produces a final repository state that is officially resolved by the SWE-bench harness, meaning that the patched repository passes the official validation. Handoff debt is the additional successor effort induced by an insufficient handoff view. In our experiments, this effort is quantified by agent events and cumulative prompt tokens, while official validation checks whether the takeover resolves the task. During continuation, the debt can appear as repeated validation, redundant file inspection, regressions, or ambiguity about predecessor intent.

This definition of handoff debt separates resumability from raw model capability. A strong successor agent may solve a task without predecessor context, but only after reconstructing the state. A weaker successor agent may fail even with a good handoff. Our primary comparisons isolate handoff-format effects. We hold the predecessor, handoff point, repository state, and successor model fixed, varying only the handoff view.

\subsection{Handoff Point Detection}

We derive handoff points deterministically from observable predecessor events rather than from the official solution. A handoff point is eligible only if it has a checkpointed repository state, an event boundary, and precomputed handoff-view records (defined in Section~\ref{sec:handoff-views}). These eligibility rules keep the protocol implementable using only logs available at handoff time.

The current handoff-point types are:

\begin{description}
    \item[\emph{After first source edit}] the first non-test source-code edit made by the predecessor;
    \item[\emph{After first validation result}] the first observed validation, build, lint, or test result after a source edit;
    \item[\emph{After first post-failure edit}] the first source edit made after the first observed failed validation result
\end{description}

Not all predecessor runs fail validation before finishing or timing out. In our selected pool, 31 of 75 source tasks contain an eligible post-failure-edit handoff point.

We run all detected handoff points for the selected source tasks rather than selecting only promising or failed states. Our comparisons are therefore over handoff points, not only over original SWE-bench issues.

\subsection{Handoff States}

We use handoff states later to separate two takeover cases that can share the same final solved label: finishing unresolved work and preserving work that is already correct. We therefore label the checkpointed repository state before takeover. We call these labels \emph{handoff states}, which describe the repository state the successor receives at a handoff point.

\paragraph{\emph{Needs completion}.}
The checkpointed repository is unresolved and a successful takeover resolves it.

\paragraph{\emph{Already solved; preserve}.}
The checkpointed repository is already resolved and a successful takeover keeps it resolved. 

\paragraph{\emph{Existing behavior broken}.}
This rare diagnostic class contains 10 instances and is used only for diagnostic breakdowns. The checkpointed repository fails tests that were passing before the predecessor’s edit, and a successful takeover must repair the task without keeping that regression.

\section{Handoff Views}
\label{sec:handoff-views}

Each takeover run starts from the same checkpointed repository $s_t$ and original task prompt. The intervention is the handoff view $h_t$, the predecessor context given to the successor. The four formats below differ only in how much of the predecessor trajectory $\tau_{\leq t}$ they expose or compress.

\paragraph{\emph{Repository only}.}
The successor receives the checkpointed repository and the original task prompt, but no explicit predecessor history. The condition is still a handoff, but it transfers only filesystem state to the successor. It measures how much of the predecessor's progress is recoverable from code alone.

\paragraph{\emph{Raw trace}.}
The successor receives the predecessor event trace up to the checkpoint, marked as historical context. This condition maximizes observable information, including commands, observations, edits, and failed attempts. It is an upper bound on available context, but not a scalable documentation strategy.

\paragraph{\emph{Summary notes}.}
The successor receives natural-language notes generated from the predecessor event timeline before the checkpoint. A predecessor agent generates these notes from the event log before takeover begins and the successor does not participate in writing them. This format tests whether free-form compression can preserve enough intent and evidence without exposing every event.

\paragraph{\emph{Structured notes}.}
The successor receives a concise structured handoff note that records the information needed to resume the task using predefined fields. Some fields are filled deterministically from checkpoint metadata and logs, including changed source files, non-source artifacts, latest source change, latest validation command, and handoff state. The predecessor agent completes the remaining fields using only predecessor-observable evidence such as the original issue, event logs, command outputs, and checkpoint repository state. 

Takeover prompts present handoff text as historical evidence rather than ground truth and instruct successors to inspect and verify the repository before relying on it. In Appendix~\ref{app:reproducibility}, we provide the note-generation settings, while Appendix~\ref{app:prompts} presents the prompt templates, full schema, and an example structured handoff excerpt.

\section{Experimental Design}

\subsection{Source Tasks and Handoff Tasks}

We start from SWE-bench Verified tasks \citep{jimenez2024swebench,openai2024swebenchverified}. We retain the 15\,minutes--1\,hour and 1--4\,hour difficulty tiers, create a fixed random order with seed 20260430, and use the first 75 source tasks. These tiers are long enough to produce meaningful handoff points while keeping full takeover evaluation feasible. The takeover benchmark is larger than the source-task count because each predecessor trajectory can yield multiple deterministic handoff points, and each handoff point is evaluated under four views. The selected-75 source pool yields 181 handoff-point tasks and 724 takeover runs per successor across four views, or 2,172 takeover runs across the three successor models. The resulting benchmark construction is summarized in Table~\ref{tab:benchmark-construction}.

\begin{table}[t]
\centering
\footnotesize
\setlength{\tabcolsep}{2pt}
\begin{tabular}{@{}>{\raggedright\arraybackslash}p{0.31\columnwidth}@{\hspace{0.02\columnwidth}}>{\raggedright\arraybackslash}p{0.43\columnwidth}r@{}}
\toprule
\textbf{Stage} & \textbf{Breakdown} & \textbf{Count} \\
\midrule
Source tasks & SWE-bench Verified & 75 \\
\addlinespace
Predecessor runs & Qwen & 75 \\
\addlinespace
Handoff points & \emph{After first source edit} & 75 \\
 & \emph{After first validation result} & 75 \\
 & \emph{After first post-failure edit} & 31 \\
 & \textbf{Total handoff points} & \textbf{181} \\
\addlinespace
Handoff states & \emph{Needs completion} & 110 \\
 & \emph{Already solved; preserve} & 61 \\
 & \emph{Existing behavior broken} & 10 \\
 & \textbf{Total state-labeled handoff points} & \textbf{181} \\
\addlinespace
Handoff views & \emph{Repository only} & 1 \\
 & \emph{Raw trace} & 1 \\
 & \emph{Summary notes} & 1 \\
 & \emph{Structured notes} & 1 \\
 & \textbf{Views per handoff} & \textbf{4} \\
\addlinespace
Takeover runs & 181 handoffs $\times$ 4 views & 724/model \\
\addlinespace
\textbf{Total takeover runs} & \textbf{724/model $\times$ 3 successors} & \textbf{2,172} \\
\bottomrule
\end{tabular}
\caption{\textbf{Construction of the takeover benchmark.} The selected-75 task pool expands into 181 handoff points and 2,172 takeover runs across the three successor models.}
\label{tab:benchmark-construction}
\end{table}

\subsection{Runtime}

We use an OpenHands-style coding-agent environment \citep{openhands2024} with terminal actions, file editing, repository freezing at handoff points, and official SWE-bench validation. Provider-native tool calling is disabled, but models still use tools through OpenHands' textual action protocol, keeping the tool interface consistent across successors while retaining a realistic full-agent runtime.

\subsection{Models}

In the main study, all handoff points come from Qwen predecessor runs. This fixes the predecessor distribution so the main intervention is the successor's handoff view. We evaluate Qwen, Gemma, and Devstral successors.\footnote{\raggedright Model pages: \href{https://huggingface.co/Qwen/Qwen3.6-27B}{\nolinkurl{Qwen/Qwen3.6-27B}}, \href{https://huggingface.co/google/gemma-4-31B-it}{\nolinkurl{google/gemma-4-31B-it}}, and \href{https://huggingface.co/mistralai/Devstral-Small-2-24B-Instruct-2512}{\nolinkurl{mistralai/Devstral-Small-2-24B-Instruct-2512}}.\par} All are served through OpenAI-compatible local endpoints, ensuring a consistent runtime protocol across successors.

We evaluate Qwen-to-Qwen, Qwen-to-Gemma, and Qwen-to-Devstral takeover pairs. Cross-model takeover tests whether handoff effects persist when the successor changes. We do not interpret these conditions as a model leaderboard because the intervention is handoff format, not model selection.

\subsection{Validation and Metrics}

We use the same scoring procedure for every handoff view. The primary outcome is
official SWE-bench resolution after takeover. The primary cost metrics are
cumulative prompt tokens and agent events, which measure how much interaction
the successor needs after receiving the handoff. We detail the runtime limits and note-generation settings in Appendix~\ref{app:reproducibility}.

\section{Results}

Context-bearing handoffs reduce rediscovery cost across all three successors. The key pattern is that context helps effort more than final resolution: it sharply reduces rediscovery work, while solved-rate gains are smaller and less uniform. A repository-only successor receives the same checkpointed files $s_t$ as every other view, but no record of what the predecessor did. Without predecessor context, the successor must reconstruct what was changed, tested, and observed.

\subsection{Rediscovery Cost}

Across all three successors and all handoff views, context-bearing handoffs reduce both agent events and prompt tokens relative to repository-only takeover at the same handoff point, as shown in Table~\ref{tab:no-handoff-vs-handoff}. \emph{Raw trace} reduces median agent events by 57--59\%, while \emph{Summary notes} and \emph{Structured notes} reduce events by 20--46\% (Figure~\ref{fig:result-event-reduction}). Prompt-token reductions are also consistent, at 42--63\%.

Repository-only takeover can start from a compact prompt, but the successor must reconstruct predecessor intent, evidence, and failure history through additional interaction. These events are not wall-clock time; reducing them by dozens per takeover means fewer repeated runtime interactions. Raw traces and notes move some of that information into the handoff, reducing the successor's rediscovery work. In Figure~\ref{fig:result-initial-prompt-size}, we report the first-prompt sizes associated with this tradeoff.

\begin{table*}[t]
\centering
\small
\setlength{\tabcolsep}{3pt}
\begin{tabular*}{\textwidth}{@{\extracolsep{\fill}}lrrrr@{}}
\toprule
\textbf{View} & \textbf{Runs} & \textbf{Solved rate ($\Delta$pp)} & \textbf{Agent events ($\Delta$\%)} & \textbf{Prompt tokens ($\Delta$\%)} \\
\midrule
\multicolumn{5}{@{}l}{\textbf{Qwen$\rightarrow$Qwen}} \\
\addlinespace[1pt]
\textit{Repository only} & 181 & 46.4\% & 99 & 1.63M \\
\addlinespace[2pt]
\multicolumn{5}{@{}l}{\textit{With predecessor context}} \\
\quad \emph{Raw trace} & 181 & 52.5\% (\textcolor{handoffgreen}{+6.1} pp) & 41 (\textcolor{handoffgreen}{-59\%}) & 811k (\textcolor{handoffgreen}{-50\%}) \\
\quad \emph{Summary notes} & 181 & 51.4\% (\textcolor{handoffgreen}{+5.0} pp) & 53 (\textcolor{handoffgreen}{-46\%}) & 602k (\textcolor{handoffgreen}{-63\%}) \\
\quad \emph{Structured notes} & 181 & 50.8\% (\textcolor{handoffgreen}{+4.4} pp) & 55 (\textcolor{handoffgreen}{-44\%}) & 660k (\textcolor{handoffgreen}{-60\%}) \\
\addlinespace[2pt]
\midrule
\multicolumn{5}{@{}l}{\textbf{Qwen$\rightarrow$Gemma}} \\
\addlinespace[1pt]
\textit{Repository only} & 181 & 42.5\% & 49 & 738k \\
\addlinespace[2pt]
\multicolumn{5}{@{}l}{\textit{With predecessor context}} \\
\quad \emph{Raw trace} & 181 & 49.2\% (\textcolor{handoffgreen}{+6.6} pp) & 21 (\textcolor{handoffgreen}{-57\%}) & 300k (\textcolor{handoffgreen}{-59\%}) \\
\quad \emph{Summary notes} & 181 & 44.2\% (\textcolor{handoffgreen}{+1.7} pp) & 33 (\textcolor{handoffgreen}{-33\%}) & 319k (\textcolor{handoffgreen}{-57\%}) \\
\quad \emph{Structured notes} & 181 & 43.6\% (\textcolor{handoffgreen}{+1.1} pp) & 39 (\textcolor{handoffgreen}{-20\%}) & 317k (\textcolor{handoffgreen}{-57\%}) \\
\addlinespace[2pt]
\midrule
\multicolumn{5}{@{}l}{\textbf{Qwen$\rightarrow$Devstral}} \\
\addlinespace[1pt]
\textit{Repository only} & 181 & 34.3\% & 175 & 3.94M \\
\addlinespace[2pt]
\multicolumn{5}{@{}l}{\textit{With predecessor context}} \\
\quad \emph{Raw trace} & 181 & 49.2\% (\textcolor{handoffgreen}{+14.9} pp) & 73 (\textcolor{handoffgreen}{-58\%}) & 1.66M (\textcolor{handoffgreen}{-58\%}) \\
\quad \emph{Summary notes} & 181 & 43.6\% (\textcolor{handoffgreen}{+9.4} pp) & 123 (\textcolor{handoffgreen}{-30\%}) & 2.30M (\textcolor{handoffgreen}{-42\%}) \\
\quad \emph{Structured notes} & 181 & 44.8\% (\textcolor{handoffgreen}{+10.5} pp) & 125 (\textcolor{handoffgreen}{-29\%}) & 2.30M (\textcolor{handoffgreen}{-42\%}) \\
\bottomrule
\end{tabular*}
\caption{\textbf{Repository-only handoff versus handoffs that include predecessor context, by successor model.} Baseline rows report absolute values; context rows report deltas in parentheses. Solved-rate deltas are percentage-point changes; cost deltas are relative changes against repository-only for the same successor.}
\label{tab:no-handoff-vs-handoff}
\end{table*}

We use a matched comparison that pairs each context-bearing run with the repository-only run from the same handoff point and successor. Each pair shares the same repository state, predecessor trajectory, and successor model, while only the handoff view changes. We then report bootstrap confidence intervals for matched run-level event reductions in Table~\ref{tab:efficiency-effects}. All intervals remain below zero, supporting that event reductions hold across matched runs rather than only in aggregate medians.

\begin{table*}[t]
\centering
\small
\setlength{\tabcolsep}{2pt}
\begin{tabular*}{\textwidth}{@{\extracolsep{\fill}}lrrrrr@{}}
\toprule
\textbf{View} &
\makecell{\textbf{Matched}\\\textbf{runs}} &
\makecell{\textbf{Repo-only}\\\textbf{agent events}} &
\makecell{\textbf{Agent events}\\\textbf{($\Delta$\%)}} &
\makecell{\textbf{95\% CI}\\\textbf{for $\Delta$ events}} &
\makecell{\textbf{Prompt tokens}\\\textbf{($\Delta$\%)}} \\
\midrule
\multicolumn{6}{@{}l}{\textbf{Qwen$\rightarrow$Qwen}} \\
\addlinespace[1pt]
\emph{Raw trace} & 181 & 99 & 41 (\textcolor{handoffgreen}{-59\%}) & [-50\%, -42\%] & 798k (\textcolor{handoffgreen}{-51\%}) \\
\emph{Summary notes} & 181 & 99 & 53 (\textcolor{handoffgreen}{-46\%}) & [-38\%, -28\%] & 572k (\textcolor{handoffgreen}{-65\%}) \\
\emph{Structured notes} & 181 & 99 & 55 (\textcolor{handoffgreen}{-44\%}) & [-34\%, -24\%] & 646k (\textcolor{handoffgreen}{-60\%}) \\
\addlinespace[4pt]
\multicolumn{6}{@{}l}{\textbf{Qwen$\rightarrow$Gemma}} \\
\addlinespace[1pt]
\emph{Raw trace} & 181 & 49 & 21 (\textcolor{handoffgreen}{-57\%}) & [-47\%, -33\%] & 300k (\textcolor{handoffgreen}{-59\%}) \\
\emph{Summary notes} & 181 & 49 & 33 (\textcolor{handoffgreen}{-33\%}) & [-25\%, -8\%] & 319k (\textcolor{handoffgreen}{-57\%}) \\
\emph{Structured notes} & 181 & 49 & 39 (\textcolor{handoffgreen}{-20\%}) & [-18\%, -1\%] & 317k (\textcolor{handoffgreen}{-57\%}) \\
\addlinespace[4pt]
\multicolumn{6}{@{}l}{\textbf{Qwen$\rightarrow$Devstral}} \\
\addlinespace[1pt]
\emph{Raw trace} & 181 & 175 & 73 (\textcolor{handoffgreen}{-58\%}) & [-45\%, -22\%] & 1.65M (\textcolor{handoffgreen}{-58\%}) \\
\emph{Summary notes} & 181 & 175 & 123 (\textcolor{handoffgreen}{-30\%}) & [-28\%, -15\%] & 2.28M (\textcolor{handoffgreen}{-42\%}) \\
\emph{Structured notes} & 181 & 175 & 125 (\textcolor{handoffgreen}{-29\%}) & [-28\%, -17\%] & 2.29M (\textcolor{handoffgreen}{-42\%}) \\
\bottomrule
\end{tabular*}
\caption{\textbf{Matched-run uncertainty analysis for the efficiency reductions} in Table~\ref{tab:no-handoff-vs-handoff}. Each context-bearing run is paired with the repository-only run from the same handoff point and successor; deltas are relative to that matched baseline. Agent event and prompt token columns report medians over matched pairs. Confidence intervals are bootstrapped over matched run-level event reductions; all intervals remain below zero.}
\label{tab:efficiency-effects}
\end{table*}

The first supplementary robustness check repeats Qwen-to-Qwen takeovers on a
stratified subset, with three attempts per handoff point. The event reductions
persist under reruns: across context-bearing views, median agent events fall by
43--59\% relative to repository-only takeover (Appendix Table~\ref{tab:takeover-sensitivity}). In
Appendix~\ref{app:robustness-selection}, we report the selection rule.

\subsection{Solved-Rate Effects}

\begin{figure*}[!t]
\centering
\includegraphics[width=\textwidth]{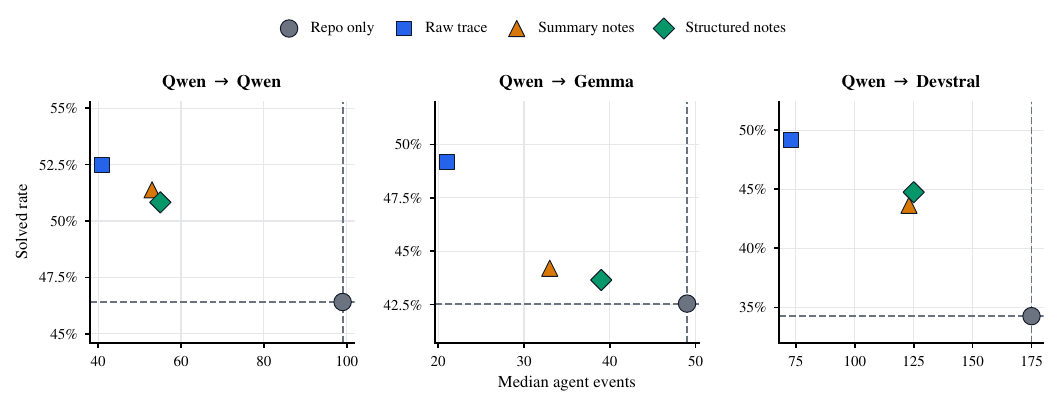}
\caption{\textbf{Solved rate versus median agent events for each successor and handoff view}. Dashed crosshairs mark the repository-only baseline within each successor condition. Axis ranges differ across panels, reflecting each model's repository-only rediscovery cost and solved-rate range.}
\label{fig:result-success-cost}
\end{figure*}

Solved-rate gains are positive but less consistent than the efficiency reductions. Matched comparisons against repository-only takeover show raw-trace gains for all successors (+6.1 to +14.9 percentage points). Note-based gains for Qwen and Gemma successors are not statistically significant at $\alpha=0.05$, while note-based gains for Devstral are significant (+9.4 to +10.5 points). The accuracy-effort tradeoff is shown in Figure~\ref{fig:result-success-cost}. We therefore treat solved-rate gains as supporting evidence. Predecessor context usually preserves or improves final resolution, while its most stable effect is reducing successor effort. We present the full matched-run solved-rate confidence intervals in Appendix Table~\ref{tab:solved-uncertainty}.

\subsection{Handoff View Tradeoffs}

\emph{Raw trace} is the highest-information view and often produces the strongest solved-rate improvements, especially for Devstral. It also produces much larger initial prompts, with a median of 87k characters compared with 7.2k for \emph{Repository only}, 9.8k for \emph{Summary notes}, and 10.0k for \emph{Structured notes}, before any takeover interaction begins (Appendix Figure~\ref{fig:result-initial-prompt-size}; Appendix Table~\ref{tab:artifact-size}). The larger initial context can still reduce total prompt tokens, since the successor needs fewer exploratory turns later.

\subsection{Where Handoff Debt Is Largest}

Handoff debt is not uniform across handoff-point types. At the \emph{After first post-failure edit} handoff point, the predecessor has accumulated failure evidence, including a validation result, a changed file, and a first response to negative feedback. A successor seeing only repository files cannot tell what was tested, what failed, or why the file changed. Repository-only successors at these handoff points require the most interaction, with median costs from 122 agent events for Qwen-to-Qwen to 191 for Qwen-to-Devstral (Appendix Table~\ref{tab:checkpoint-kind}). Context-bearing handoffs also show large solved-rate gains here, reaching +12.9 percentage points for Qwen-to-Qwen and +19.4 points for Qwen-to-Devstral.

We also group handoff points by handoff state, meaning what kind of repository state the successor receives. Among the 110 \emph{Needs completion} handoff points, repository-only successors resolve only 13.6--23.6\% of tasks (Appendix Table~\ref{tab:checkpoint-class}). Context-bearing views raise completion rates across all three successors, with the best view reaching 23.6--28.2\%. These gains matter most here because \emph{Needs completion} is the core handoff scenario, where a predecessor stopped before finishing and a successor must carry the work forward.

\subsection{Cross-Model Checks}

Across Qwen-to-Qwen, Qwen-to-Gemma, and Qwen-to-Devstral takeovers, the main pattern is stable, but the preferred handoff format differs by successor. Raw traces, summaries, and structured notes do not produce one universal ranking. The cross-model comparison therefore supports the rediscovery-cost result while showing that resumability depends on the successor as well as the handoff artifact.

The second supplementary robustness check asks whether the pattern depends on which model generates the handoff, rather than which successor model receives it.
We run Qwen, Gemma, and Devstral predecessors on the same selected-75
source-task pool used in the main experiment, and evaluate the resulting handoff
points with all three successors. This yields 510 handoff points and 6,120
takeover runs.
The three predecessors leave different handoff patterns, summarized in
Appendix Table~\ref{tab:predecessor-patterns}. The efficiency pattern still
holds across these predecessor families. In all nine predecessor--successor
pairs, raw-trace takeover requires fewer median agent events than repository-only
takeover, and note-based views recover most of that saving while keeping the
handoff artifact bounded. Appendix Table~\ref{tab:predecessor-robustness}
reports the fixed-Devstral-successor slice in detail.
\section{Analysis}
\label{sec:analysis}

\subsection{Rediscovery and Evidence}

Without predecessor context, successors reconstruct intent from the repository state alone, inspecting changed files, rerunning checks, and re-deriving what was attempted. Repository-only takeover receives the files, but the successor must infer why they changed, what evidence was observed, and which assumptions are still valid. Repository-only runs can therefore have compact initial prompts while still accumulating more model actions, tool interactions, and prompt tokens over the full takeover. The debt is paid during reconstruction.

The value of context-bearing handoffs depends on the type of evidence available at the handoff point. At \emph{After first source edit}, the successor receives a patch but little feedback. At \emph{After first validation result} and \emph{After first post-failure edit}, the handoff point often contains failure evidence, including what was checked, what failed, and what changed in response. Context-bearing handoffs are most useful here because they pass along evidence the successor would otherwise have to reconstruct. For diagnostic reference, Appendix Tables~\ref{tab:checkpoint-kind} and~\ref{tab:checkpoint-class} report the handoff-point and handoff-state breakdowns separately.

\subsection{Choosing a Handoff View}

The four formats trade off how much predecessor evidence they transfer against how much that evidence burdens the successor's initial context. \emph{Raw trace} reduces rediscovery because it exposes commands, observations, and failed attempts directly, but it is large, noisy, and unbounded. Even in our takeover experiments, its first prompts are far larger than those of note-based handoffs (Appendix Figure~\ref{fig:result-initial-prompt-size}). In longer-running work, replaying the full event history becomes impractical and makes it harder to control what information is passed forward.

This contrast reflects how costs are distributed across the takeover. \emph{Raw trace} starts with a large prompt, but the successor often needs fewer exploratory turns later, reducing total token use. Repository-only takeover appears cheap at handoff time, but the successor pays later by repeating tests and re-deriving intent. Handoff debt is therefore measured over the whole takeover, not only by the first prompt.

\emph{Summary notes} are competitive in several experimental conditions, showing that free-form compression can preserve useful state on benchmark-scale tasks. \emph{Structured notes} expose a different tradeoff. They make the handoff a bounded continuation contract rather than a free-form narrative. A structured record contains the same continuation fields for each takeover, including what changed, what evidence was observed, what remains uncertain, what should be verified, and what may need rollback. That regular structure makes the handoff easier to inspect, filter, and reuse.

Context can still hurt when compressed notes omit crucial evidence or when a successor over-trusts the predecessor's interpretation. For Gemma note-based handoffs in the \emph{Needs completion} state, we observe small negative solved-rate deltas (Appendix Table~\ref{tab:checkpoint-class}). These cases motivate verification of handoff text rather than blind trust. 
The observed failure modes are summarized in Section~\ref{app:failure-modes}. They also explain why \emph{Needs completion} and \emph{Already solved; preserve} handoff points should be separated in diagnostics. Taking over solved work and taking over unresolved work are different continuation problems.

The same missing context can affect successors differently. For some successors, missing context mainly increases effort; for others, missing evidence becomes a final-resolution loss. Resumability is therefore a joint property of predecessor state, handoff artifact, and successor behavior. \emph{Structured notes} are most useful at handoff points with validation evidence, especially \emph{After first validation result} and \emph{After first post-failure edit}, where the repository alone cannot convey what the predecessor tested, what failed, and how it responded.

\section{Related Work}

\paragraph{Human handoff and coordination.}
Human software teams use tickets, reviews, commit messages, and design notes to make work resumable. Software-engineering research has long studied how coordination and distributed knowledge shape software work \citep{brooks1987nosilverbullet,herbsleb2003empirical}. Structured handoff notes are inspired by this practice, but our recipient is another coding agent, and outcomes are measured by official validation plus continuation cost.

\paragraph{Coding-agent evaluation.}
SWE-bench evaluates whether language models can resolve real GitHub issues by editing repository code and passing held-out tests \citep{jimenez2024swebench}; SWE-bench Verified refines that task set for more reliable scoring \citep{openai2024swebenchverified}. SWE-agent, Agentless, and OpenHands show how interfaces and full-agent runtimes affect software-agent performance \citep{yang2024sweagent,xia2024agentless,openhands2024}. ContextBench adds process-oriented metrics for repository-context use \citep{li2026contextbench}. Our focus is complementary. Given an interrupted trajectory, we measure whether the resulting state is resumable by a separate successor.

\paragraph{Agent state and tool use.}
AgentBench, MINT, WebArena, and $\tau$-bench evaluate multi-turn tool use and simulated user interaction \citep{liu2023agentbench,wang2023mint,zhou2023webarena,yao2024taubench}. ReAct, Toolformer, and ToolLLM show that external actions and observations reshape model behavior \citep{yao2023react,schick2023toolformer,qin2023toolllm}; Reflexion and Voyager use feedback or self-generated programs to improve later behavior \citep{shinn2023reflexion,wang2023voyager}. Multi-agent systems such as AutoGen, LangGraph, CrewAI, and Agyn make state transfer operationally important \citep{wu2023autogen,langgraph2024,crewai2024,benkovich2026agyn}, but they do not isolate the handoff artifact itself by holding the repository checkpoint fixed and varying only the continuation context.

\paragraph{Agent memory and summarization.}
Long-running agents rely on memory compression, trajectory summarization, and context management. MemGPT treats context as managed memory; StreamingLLM and Longformer show that long context is also a systems and architecture constraint \citep{packer2023memgpt,xiao2023streamingllm,beltagy2020longformer}. Our handoff views differ from generic memory compression because the compressed state is consumed by a potentially different successor and must support validated continuation from a concrete repository state. We use fixed handoff views to isolate the handoff artifact, but the same takeover protocol could evaluate adaptive memory policies by holding the checkpoint fixed and comparing the artifact they produce.

\paragraph{Summaries and repair evidence.}
Handoff notes are related to code summarization and developer documentation, but their purpose is operational rather than descriptive. Prior work studies natural-language summaries of source code, comments, and changes \citep{haiduc2010summarizing,iyer2016summarizing}. Handoff is also related to repair systems that use history or evidence. HAFixAgent injects curated repository history into agentic automated program repair, while REFINE and TraceRepair use patch, review, or execution evidence to improve repair loops \citep{shi2025hafixagent,pabba2025refine,wu2026tracerepair}. We instead measure whether predecessor-generated evidence helps a separate successor continue from an interrupted repository state.

\section{Conclusion}

We introduced handoff debt as a way to evaluate whether another agent can take over partially completed work correctly and efficiently. In SWE-bench Verified takeover experiments, repository state alone often leaves successors to rediscover predecessor context, producing substantially more agent events and prompt tokens. \emph{Raw trace} is informative but unbounded; compact handoff artifacts offer a practical middle ground. Coding-agent evaluation should therefore measure whether agent work remains understandable, verifiable, and resumable, not only whether it is eventually finished.

The findings suggest two practical implications. First, benchmarks should report handoff debt alongside final resolution, asking whether another agent can understand, verify, and continue the work without paying a large rediscovery cost. Second, agent systems should make handoff output a deliberate part of the workflow, not merely a log produced when a run stops.

\section*{Limitations}

\paragraph{Runtime scope.}
Our conclusions are within an OpenHands-style coding-agent runtime. We fixed one runtime to keep the comparison controlled. We chose OpenHands because it models real coding-agent work and produces the trajectory evidence our protocol needs, including file edits, terminal output, validation results, and multi-turn tool interactions. Other coding-agent runtimes may change the absolute event counts, token counts, and solved rates. But the rediscovery problem itself is not specific to OpenHands. In any runtime or coding-agent framework, the successor resumes from a frozen repository plus whatever predecessor context the handoff provides, and must reconstruct the prior work that is not visible from that handoff artifact. We therefore expect the same ordering across runtimes: context-bearing handoffs reduce rediscovery relative to the frozen repository alone, with the magnitude varying by runtime.

\paragraph{Single-run handoffs.}
Each handoff point is evaluated once per handoff view in the main study rather than with repeated independent attempts. A robustness study with three attempts on each of 40 stratified handoff points shows the same pattern. Context-bearing views require 43--59\% fewer agent events across repeated runs (Table~\ref{tab:takeover-sensitivity}). This suggests that the primary efficiency result is not an artifact of one attempt per handoff view.

\paragraph{Source task pool.}
We use 75 tasks from SWE-bench Verified as source tasks. The handoff benchmark is substantially larger because each source task yields multiple deterministic handoff points, producing 181 handoff-point tasks and 2,172 total runs across all views and successors.

\paragraph{Validation coverage.}
We use official SWE-bench validation, which scores whether the submitted patch passes the held-out tests. This covers patch correctness but not maintainability or broader human usefulness. For our primary question this is sufficient because test passage provides a common resolution measure while agent events and prompt tokens quantify rediscovery cost.

\section*{Ethical Considerations}

This work uses public SWE-bench Verified tasks and generated agent trajectories. It does not involve human-subject data or private user data, and the results should be read as benchmark evidence about validation and continuation cost rather than as a guarantee of deployment safety.

\bibliography{references}

\appendix
\section{Diagnostic Breakdowns}
\label{app:diagnostics}

These tables provide additional views of the same selected-75 source-task takeover runs reported in the main paper. They are diagnostic rather than central to the main claim.

\begin{table*}[t]
\centering
\small
\setlength{\tabcolsep}{3pt}
\begin{tabular*}{\textwidth}{@{\extracolsep{\fill}}lrrr@{}}
\toprule
\textbf{View} &
\textbf{Runs} &
\makecell{\textbf{Agent events}\\\textbf{($\Delta$\%)}} &
\makecell{\textbf{Prompt tokens}\\\textbf{($\Delta$\%)}} \\
\midrule
\multicolumn{4}{@{}l}{\textbf{Qwen$\rightarrow$Qwen repeated takeovers: 40 handoff points $\times$ 3 repeats}} \\
\addlinespace[1pt]
\textit{Repository only} & 120 & 102.5 & 1.62M \\
\addlinespace[2pt]
\multicolumn{4}{@{}l}{\textit{With predecessor context}} \\
\quad \emph{Raw trace} & 120 & 42 (\textcolor{handoffgreen}{-59\%}) & 906k (\textcolor{handoffgreen}{-44\%}) \\
\quad \emph{Summary notes} & 120 & 52 (\textcolor{handoffgreen}{-49\%}) & 591k (\textcolor{handoffgreen}{-64\%}) \\
\quad \emph{Structured notes} & 120 & 58 (\textcolor{handoffgreen}{-43\%}) & 677k (\textcolor{handoffgreen}{-58\%}) \\
\bottomrule
\end{tabular*}
\caption{\textbf{Repeated-takeover sensitivity} on 40 Qwen-to-Qwen handoff points. Each view contains 120 takeover runs: three repeated takeovers for each handoff point. Agent events and prompt tokens report medians over takeover runs; cost parentheses report relative changes against repository-only.}
\label{tab:takeover-sensitivity}
\end{table*}

\subsection{Robustness-Study Selection}
\label{app:robustness-selection}

\textbf{Repeated-takeover sensitivity.}
This study uses 40 Qwen-to-Qwen handoff points selected from the main
selected-75 pool: 30 \emph{Needs completion} handoff points, balanced across the
three handoff-point types, and 10 \emph{Already solved; preserve} handoff points
split across \emph{After first source edit} and \emph{After first validation result}.
Each handoff point is evaluated under all four handoff views with three attempts,
yielding 120 takeover runs per view.

\textbf{Predecessor robustness.}
This study uses the same selected-75 source-task pool as the main study. We run
Qwen, Gemma, and Devstral as predecessors and evaluate each eligible handoff
point under all four handoff views with all three successors. This gives 510
handoff points and 6,120 takeover runs. 

Table~\ref{tab:predecessor-patterns}
summarizes how the three predecessors differ before takeover. Gemma produces shorter traces and reaches validation less often. Devstral
produces longer traces, with 3.4$\times$ Gemma's median events and the most
handoff points after a post-failure edit. Qwen sits between them. These
differences come from the model runs themselves; we did not instruct the models
to behave this way.

Table~\ref{tab:predecessor-robustness}
shows the Devstral-successor setting in detail; the same efficiency pattern
holds across all successor models.

\begin{table*}[t]
\centering
\small
\setlength{\tabcolsep}{3pt}
\begin{tabular*}{\textwidth}{@{\extracolsep{\fill}}lrrrrrr@{}}
\toprule
\textbf{Predecessor} &
\makecell{\textbf{Handoff}\\\textbf{points}} &
\makecell{\textbf{Median}\\\textbf{events}} &
\makecell{\textbf{Median}\\\textbf{prompt}\\\textbf{tokens}} &
\makecell{\textbf{After first}\\\textbf{source edit}} &
\makecell{\textbf{After first}\\\textbf{validation result}} &
\makecell{\textbf{After first}\\\textbf{post-failure edit}} \\
\midrule
Qwen & 181 & 107 & 1.83M & 75 & 75 & 31 \\
Gemma & 145 & 49 & 0.73M & 73 & 58 & 14 \\
Devstral & 184 & 167 & 3.78M & 72 & 72 & 40 \\
\bottomrule
\end{tabular*}
\caption{\textbf{Predecessor handoff patterns.} Counts are generated from the
same selected-75 source-task pool. The last three columns count eligible
handoff points generated after each interruption rule.}
\label{tab:predecessor-patterns}
\end{table*}

\begin{table*}[t]
\centering
\small
\setlength{\tabcolsep}{3pt}
\begin{tabular*}{\textwidth}{@{\extracolsep{\fill}}lrrr@{}}
\toprule
\textbf{View} &
\textbf{Runs} &
\makecell{\textbf{Agent events}\\\textbf{($\Delta$\%)}} &
\makecell{\textbf{Prompt tokens}\\\textbf{($\Delta$\%)}} \\
\midrule
\multicolumn{4}{@{}l}{\textbf{Qwen$\rightarrow$Devstral}} \\
\addlinespace[1pt]
\textit{Repository only} & 181 & 175 & 3.94M \\
\addlinespace[2pt]
\multicolumn{4}{@{}l}{\textit{With predecessor context}} \\
\quad \emph{Raw trace} & 181 & 73 (\textcolor{handoffgreen}{-58\%}) & 1.66M (\textcolor{handoffgreen}{-58\%}) \\
\quad \emph{Summary notes} & 181 & 123 (\textcolor{handoffgreen}{-30\%}) & 2.30M (\textcolor{handoffgreen}{-42\%}) \\
\quad \emph{Structured notes} & 181 & 125 (\textcolor{handoffgreen}{-29\%}) & 2.30M (\textcolor{handoffgreen}{-42\%}) \\
\midrule
\multicolumn{4}{@{}l}{\textbf{Gemma$\rightarrow$Devstral}} \\
\addlinespace[1pt]
\textit{Repository only} & 145 & 179 & 4.15M \\
\addlinespace[2pt]
\multicolumn{4}{@{}l}{\textit{With predecessor context}} \\
\quad \emph{Raw trace} & 145 & 97 (\textcolor{handoffgreen}{-46\%}) & 2.28M (\textcolor{handoffgreen}{-45\%}) \\
\quad \emph{Summary notes} & 145 & 141 (\textcolor{handoffgreen}{-21\%}) & 2.73M (\textcolor{handoffgreen}{-34\%}) \\
\quad \emph{Structured notes} & 145 & 131 (\textcolor{handoffgreen}{-27\%}) & 2.63M (\textcolor{handoffgreen}{-37\%}) \\
\midrule
\multicolumn{4}{@{}l}{\textbf{Devstral$\rightarrow$Devstral}} \\
\addlinespace[1pt]
\textit{Repository only} & 184 & 173 & 4.17M \\
\addlinespace[2pt]
\multicolumn{4}{@{}l}{\textit{With predecessor context}} \\
\quad \emph{Raw trace} & 184 & 95 (\textcolor{handoffgreen}{-45\%}) & 2.76M (\textcolor{handoffgreen}{-34\%}) \\
\quad \emph{Summary notes} & 184 & 133 (\textcolor{handoffgreen}{-23\%}) & 2.51M (\textcolor{handoffgreen}{-40\%}) \\
\quad \emph{Structured notes} & 184 & 137 (\textcolor{handoffgreen}{-21\%}) & 2.76M (\textcolor{handoffgreen}{-34\%}) \\
\bottomrule
\end{tabular*}
\caption{\textbf{Predecessor-robustness sensitivity.} Each predecessor model contributes all eligible handoff points from the selected-75 source-task pool, evaluated under four handoff views with a fixed Devstral successor. Agent events and prompt tokens report medians over takeover runs; cost parentheses report relative changes against repository-only for the same predecessor.}
\label{tab:predecessor-robustness}
\end{table*}

\begin{table*}[t]
\centering
\small
\setlength{\tabcolsep}{4pt}
\begin{tabular*}{\textwidth}{@{\extracolsep{\fill}}lrrrrrr@{}}
\toprule
\textbf{View} &
\textbf{Runs} &
\makecell{\textbf{$\Delta$ solved}\\\textbf{pp}} &
\makecell{\textbf{95\%}\\\textbf{CI}} &
\makecell{\textbf{Context-only}\\\textbf{solves}} &
\makecell{\textbf{Repo-only}\\\textbf{solves}} &
\textbf{$p$} \\
\midrule
\multicolumn{7}{@{}l}{\textbf{Qwen$\rightarrow$Qwen}} \\
\addlinespace[1pt]
\emph{Raw trace} & 181 & +6.1 & [1.1, 11.0] & 17 & 6 & 0.035 \\
\emph{Summary notes} & 181 & +5.0 & [0.0, 9.9] & 16 & 7 & 0.093 \\
\emph{Structured notes} & 181 & +4.4 & [-0.6, 9.4] & 15 & 7 & 0.134 \\
\addlinespace[4pt]
\multicolumn{7}{@{}l}{\textbf{Qwen$\rightarrow$Gemma}} \\
\addlinespace[1pt]
\emph{Raw trace} & 181 & +6.6 & [1.1, 12.2] & 19 & 7 & 0.029 \\
\emph{Summary notes} & 181 & +1.7 & [-3.9, 6.6] & 13 & 10 & 0.678 \\
\emph{Structured notes} & 181 & +1.1 & [-3.3, 5.5] & 10 & 8 & 0.815 \\
\addlinespace[4pt]
\multicolumn{7}{@{}l}{\textbf{Qwen$\rightarrow$Devstral}} \\
\addlinespace[1pt]
\emph{Raw trace} & 181 & +14.9 & [8.8, 21.0] & 33 & 6 & $<$.001 \\
\emph{Summary notes} & 181 & +9.4 & [3.3, 15.5] & 25 & 8 & 0.005 \\
\emph{Structured notes} & 181 & +10.5 & [5.0, 16.0] & 23 & 4 & $<$.001 \\
\bottomrule
\end{tabular*}
\caption{\textbf{Solved-rate uncertainty against repository-only takeover.} Each comparison uses matched runs from the same handoff point for the successor model. Confidence intervals are nonparametric bootstrap intervals over matched runs. The $p$ column reports the McNemar test p-value for the discordant pairs. The final columns report the McNemar discordant pairs: context-only solves are runs solved by the context view but not repository-only, and repo-only solves are the reverse.}
\label{tab:solved-uncertainty}
\end{table*}

\begin{table*}[t]
\centering
\small
\setlength{\tabcolsep}{5pt}
\begin{tabular*}{\textwidth}{@{\extracolsep{\fill}}lrrrr@{}}
\toprule
\textbf{View} & \textbf{Runs} & \textbf{Median} & \textbf{P90} & \textbf{Max} \\
\midrule
\emph{Repository only} & 543 & 7.2k & 10k & 30k \\
\emph{Raw trace} & 543 & 87k & 150k & 471k \\
\emph{Summary notes} & 543 & 9.8k & 13k & 32k \\
\emph{Structured notes} & 543 & 10.0k & 13k & 33k \\
\bottomrule
\end{tabular*}
\caption{\textbf{Rendered first-prompt size by handoff view}, aggregated over all successor conditions. Size columns report prompt characters. \emph{Raw trace} carries substantially more initial context than repository-only or note-based views, before any takeover interaction begins.}
\label{tab:artifact-size}
\end{table*}

\begin{figure*}[t]
\centering
\includegraphics[width=\textwidth]{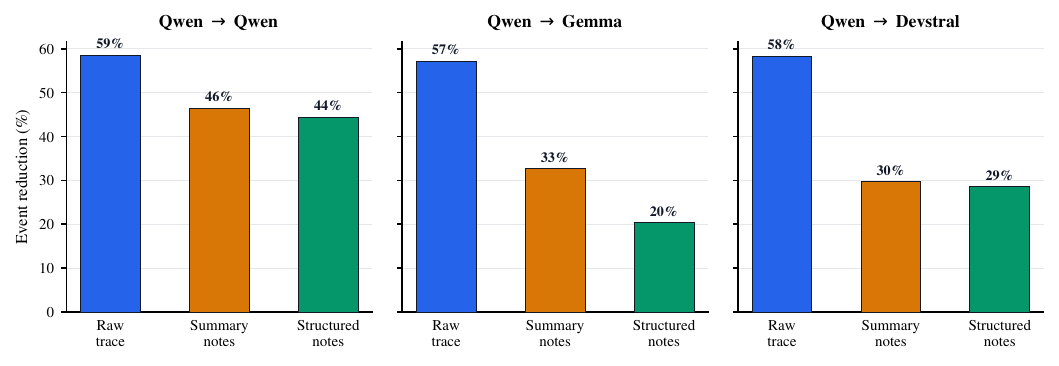}
\caption{\textbf{Reduction in median agent events relative to repository-only takeover.} Context-bearing handoffs consistently reduce rediscovery effort across successor models. Repository-only baselines are 99, 49, and 175 median agent events per takeover run for Qwen-to-Qwen, Qwen-to-Gemma, and Qwen-to-Devstral, respectively.}
\label{fig:result-event-reduction}
\end{figure*}

\begin{figure*}[t]
\centering
\includegraphics[width=0.72\textwidth]{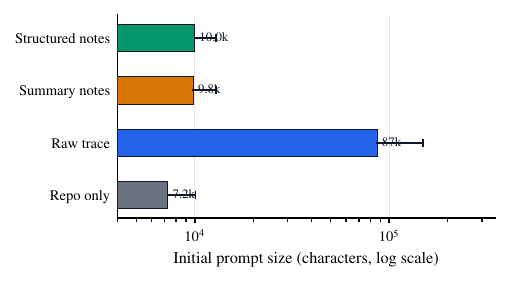}
\caption{\textbf{Log-scale chart of rendered first-prompt size by handoff view}. Bars show median initial prompt characters and whiskers show the 90th percentile; the pattern is consistent across successors. \emph{Raw trace} is much larger than repository-only and note-based handoffs before the successor takes any action.}
\label{fig:result-initial-prompt-size}
\end{figure*}

\begin{table*}[t]
\centering
\footnotesize
\setlength{\tabcolsep}{2pt}
\begin{tabular*}{\textwidth}{@{\extracolsep{\fill}}lrrrrrr@{}}
\toprule
\makecell{\textbf{Handoff}\\\textbf{point}} &
\textbf{Runs} &
\textbf{Repository only} &
\textbf{Raw trace} &
\textbf{Summary notes} &
\textbf{Structured notes} &
\makecell{\textbf{Repo-only}\\\textbf{agent events}} \\
\midrule
\multicolumn{7}{@{}l}{\textbf{Qwen$\rightarrow$Qwen}} \\
\addlinespace[1pt]
\emph{After first source edit} & 75 & 49.3\% & 50.7\% (\textcolor{handoffgreen}{+1.3}) & 52.0\% (\textcolor{handoffgreen}{+2.7}) & 49.3\% (\textcolor{handoffgray}{0.0}) & 101 \\
\emph{After first validation result} & 75 & 48.0\% & 56.0\% (\textcolor{handoffgreen}{+8.0}) & 53.3\% (\textcolor{handoffgreen}{+5.3}) & 54.7\% (\textcolor{handoffgreen}{+6.7}) & 93 \\
\emph{After first post-failure edit} & 31 & 35.5\% & 48.4\% (\textcolor{handoffgreen}{+12.9}) & 45.2\% (\textcolor{handoffgreen}{+9.7}) & 45.2\% (\textcolor{handoffgreen}{+9.7}) & 122 \\
\addlinespace[4pt]
\multicolumn{7}{@{}l}{\textbf{Qwen$\rightarrow$Gemma}} \\
\addlinespace[1pt]
\emph{After first source edit} & 75 & 41.3\% & 48.0\% (\textcolor{handoffgreen}{+6.7}) & 48.0\% (\textcolor{handoffgreen}{+6.7}) & 40.0\% (\textcolor{handoffred}{-1.3}) & 49 \\
\emph{After first validation result} & 75 & 45.3\% & 52.0\% (\textcolor{handoffgreen}{+6.7}) & 45.3\% (\textcolor{handoffgray}{0.0}) & 49.3\% (\textcolor{handoffgreen}{+4.0}) & 43 \\
\emph{After first post-failure edit} & 31 & 38.7\% & 45.2\% (\textcolor{handoffgreen}{+6.5}) & 32.3\% (\textcolor{handoffred}{-6.5}) & 38.7\% (\textcolor{handoffgray}{0.0}) & 63 \\
\addlinespace[4pt]
\multicolumn{7}{@{}l}{\textbf{Qwen$\rightarrow$Devstral}} \\
\addlinespace[1pt]
\emph{After first source edit} & 75 & 38.7\% & 49.3\% (\textcolor{handoffgreen}{+10.7}) & 44.0\% (\textcolor{handoffgreen}{+5.3}) & 42.7\% (\textcolor{handoffgreen}{+4.0}) & 169 \\
\emph{After first validation result} & 75 & 36.0\% & 53.3\% (\textcolor{handoffgreen}{+17.3}) & 45.3\% (\textcolor{handoffgreen}{+9.3}) & 50.7\% (\textcolor{handoffgreen}{+14.7}) & 171 \\
\emph{After first post-failure edit} & 31 & 19.4\% & 38.7\% (\textcolor{handoffgreen}{+19.4}) & 38.7\% (\textcolor{handoffgreen}{+19.4}) & 35.5\% (\textcolor{handoffgreen}{+16.1}) & 191 \\
\bottomrule
\end{tabular*}
\caption{\textbf{Solved-rate breakdown by handoff point}. Deltas are percentage-point changes relative to repository-only for the same successor and handoff point type. Run counts are per view. The final column shows the median repository-only agent events, a direct measure of rediscovery effort.}
\label{tab:checkpoint-kind}
\vspace{-0.6em}
\end{table*}

\begin{table*}[t]
\centering
\small
\setlength{\tabcolsep}{5pt}
\begin{tabular*}{\textwidth}{@{\extracolsep{\fill}}lrrrrr@{}}
\toprule
\textbf{Handoff state} &
\textbf{Runs} &
\textbf{Repository only} &
\textbf{Raw trace} &
\textbf{Summary notes} &
\textbf{Structured notes} \\
\midrule
\multicolumn{6}{@{}l}{\textbf{Qwen$\rightarrow$Qwen}} \\
\addlinespace[1pt]
\emph{Needs completion} & 110 & 23.6\% & 27.3\% (\textcolor{handoffgreen}{+3.6}) & 28.2\% (\textcolor{handoffgreen}{+4.5}) & 26.4\% (\textcolor{handoffgreen}{+2.7}) \\
\emph{Already solved; preserve} & 61 & 88.5\% & 93.4\% (\textcolor{handoffgreen}{+4.9}) & 91.8\% (\textcolor{handoffgreen}{+3.3}) & 95.1\% (\textcolor{handoffgreen}{+6.6}) \\
\emph{Existing behavior broken} & 10 & 40.0\% & 80.0\% (\textcolor{handoffgreen}{+40.0}) & 60.0\% (\textcolor{handoffgreen}{+20.0}) & 50.0\% (\textcolor{handoffgreen}{+10.0}) \\
\addlinespace[4pt]
\multicolumn{6}{@{}l}{\textbf{Qwen$\rightarrow$Gemma}} \\
\addlinespace[1pt]
\emph{Needs completion} & 110 & 17.3\% & 23.6\% (\textcolor{handoffgreen}{+6.4}) & 15.5\% (\textcolor{handoffred}{-1.8}) & 16.4\% (\textcolor{handoffred}{-0.9}) \\
\emph{Already solved; preserve} & 61 & 95.1\% & 96.7\% (\textcolor{handoffgreen}{+1.6}) & 98.4\% (\textcolor{handoffgreen}{+3.3}) & 98.4\% (\textcolor{handoffgreen}{+3.3}) \\
\emph{Existing behavior broken} & 10 & 0.0\% & 40.0\% (\textcolor{handoffgreen}{+40.0}) & 30.0\% (\textcolor{handoffgreen}{+30.0}) & 10.0\% (\textcolor{handoffgreen}{+10.0}) \\
\addlinespace[4pt]
\multicolumn{6}{@{}l}{\textbf{Qwen$\rightarrow$Devstral}} \\
\addlinespace[1pt]
\emph{Needs completion} & 110 & 13.6\% & 24.5\% (\textcolor{handoffgreen}{+10.9}) & 16.4\% (\textcolor{handoffgreen}{+2.7}) & 20.9\% (\textcolor{handoffgreen}{+7.3}) \\
\emph{Already solved; preserve} & 61 & 77.0\% & 95.1\% (\textcolor{handoffgreen}{+18.0}) & 98.4\% (\textcolor{handoffgreen}{+21.3}) & 93.4\% (\textcolor{handoffgreen}{+16.4}) \\
\emph{Existing behavior broken} & 10 & 0.0\% & 40.0\% (\textcolor{handoffgreen}{+40.0}) & 10.0\% (\textcolor{handoffgreen}{+10.0}) & 10.0\% (\textcolor{handoffgreen}{+10.0}) \\
\bottomrule
\end{tabular*}
\caption{\textbf{Official solved outcomes by handoff state.} Each cell reports the official solved rate for that view and state. Run counts are per view within the state. Parenthesized deltas are percentage-point changes relative to repository-only for the same successor and handoff state. The \emph{Existing behavior broken} row has only 10 instances and should be interpreted as diagnostic.}
\label{tab:checkpoint-class}
\end{table*}

\clearpage

\clearpage

\twocolumn
\section{Reproducibility Details}
\label{app:reproducibility}

\paragraph{Runtime.}
All reported model conditions use local OpenAI-compatible endpoints through the
same OpenHands-style runtime. Provider-native tool calling is disabled; models
still use tools through OpenHands' textual action protocol. Experiments were
run on NVIDIA RTX PRO 6000 Blackwell Max-Q Workstation Edition GPUs using local vLLM model
servers. The model servers use the same serving pattern across conditions:
\texttt{--max-num-seqs 16}, \texttt{--gpu-memory-utilization 0.95},
\texttt{--dtype auto}, and \texttt{--language-model-only}. We do not force a
shared context length beyond each model's respective serving configuration.
Takeover runs use a 4-hour conversation timeout and a cap of 500 agent steps.
The canonical model identifiers are
\href{https://huggingface.co/Qwen/Qwen3.6-27B}{\nolinkurl{Qwen/Qwen3.6-27B}},
\href{https://huggingface.co/google/gemma-4-31B-it}{\nolinkurl{google/gemma-4-31B-it}},
and
\href{https://huggingface.co/mistralai/Devstral-Small-2-24B-Instruct-2512}{\nolinkurl{mistralai/Devstral-Small-2-24B-Instruct-2512}}.

\paragraph{Handoff generation.}
Before takeover, the predecessor-side model writes the free-form summary notes
and the model-filled fields of structured notes. The main study uses Qwen for
this step; the predecessor-robustness study uses the corresponding predecessor
model. The calls use temperature 0 with a 1600-token cap. The remaining
structured-note fields are filled deterministically from checkpoint metadata and
logs. Note-generation cost is not included in successor agent-event or
prompt-token metrics, which measure takeover effort after the handoff is
presented. Per-run output files log prompt tokens, completion tokens, and wall-clock
time. We report prompt tokens and agent events as the primary cost metrics
because they directly measure context consumption and rediscovery effort after
takeover. Completion tokens and wall-clock time are retained for
reproducibility, but we do not use wall-clock time or GPU-hours for cross-view
comparisons because they depend on local serving load, Docker scheduling, and
parallel worker contention.

\paragraph{Statistical tests.}
For statistical uncertainty, we use 95\% nonparametric percentile bootstrap
intervals with 5,000 resamples and fixed seed 20260518. Solved-rate intervals
resample matched binary run deltas, while efficiency intervals resample matched
run-level relative event reductions. McNemar $p$-values are exact two-sided
binomial tests over discordant pairs. We do not apply a multiple-comparison
correction; the tests characterize paired effects by handoff view and successor
rather than selecting a single winning condition.

\paragraph{Task selection.}
The selected source-task order is fixed with random seed 20260430 after
filtering SWE-bench Verified to the 15 minute--1 hour and 1--4 hour difficulty
tiers. The main study contains 75 source tasks, 181 deterministic handoff
points, four handoff views, and three successor models, for 2,172 main takeover
runs. Repeated-takeover sensitivity uses 40 Qwen-to-Qwen handoff points with
three repeated takeovers per view. Predecessor robustness uses the same
selected-75 source-task pool as the main study. Qwen, Gemma, and Devstral
predecessor trajectories produce 510 eligible handoff points. Each point is
evaluated under all four handoff views with all three successors, yielding
6,120 takeover runs. Appendix Table~\ref{tab:predecessor-robustness} reports
the Devstral-successor setting in detail.
\onecolumn
\section{Prompt and Handoff Schemas}
\label{app:prompts}

We show the takeover prompt templates used in the experiments. The base task
prompt follows the OpenHands SWE-bench prompt format \citep{openhands2024} and
includes the original SWE-bench issue text, which is inserted verbatim in every
view but replaced below with \texttt{[ORIGINAL TASK PROMPT]}. Raw traces and
generated notes are abbreviated with placeholders; they are produced before
takeover from predecessor-observed events only.

\paragraph{Shared takeover instructions.}
Every takeover prompt begins with the same instruction block:

\begin{quote}\small\raggedright
\texttt{=== Takeover context ===}\\
You are continuing a coding task in a repository that already contains work
from a previous agent. The task may already be solved, partially solved, or
incorrectly solved. Inspect the current repository state before editing. Treat
handoff details as historical notes from the previous agent, not ground truth.
Treat the original task prompt as the source of truth for requirements. Verify
important claims against the current repository state before relying on them.
If the existing changes already satisfy the task, preserve them and finish once
verification is sufficient. Avoid restarting from scratch unless the current
changes are clearly wrong.
\end{quote}

\paragraph{\emph{Repository only}.}
Repository-only takeover appends only the original issue:

\begin{quote}\small\raggedright
\texttt{=== Original task prompt ===}\\
\texttt{[ORIGINAL TASK PROMPT]}\\
\texttt{=== End original task prompt ===}
\end{quote}

\paragraph{\emph{Raw trace}.}
Raw-trace takeover provides the predecessor event history before the original
issue:

\begin{quote}\small\raggedright
\texttt{=== Previous-agent raw trace ===}\\
The following raw trace is historical context from the previous agent. Use it
to understand what has been tried and what remains. Continue from the current
repository state.\\
\texttt{[EVENTS UP TO HANDOFF: step, source, event type, text]}\\
\texttt{=== End previous-agent raw trace ===}\\
\texttt{=== Original task prompt ===}\\
\texttt{[ORIGINAL TASK PROMPT]}\\
\texttt{=== End original task prompt ===}
\end{quote}

\paragraph{\emph{Summary notes}.}
Summary-note takeover inserts a concise generated summary before the original
issue:

\begin{quote}\small\raggedright
\texttt{=== Previous-agent summary notes ===}\\
Natural-language summary of the previous agent work log:\\
\texttt{[GENERATED SUMMARY: investigation, edits, validation attempts, uncertainty, next steps]}\\
\texttt{=== End previous-agent summary notes ===}\\
\texttt{=== Original task prompt ===}\\
\texttt{[ORIGINAL TASK PROMPT]}\\
\texttt{=== End original task prompt ===}
\end{quote}

\paragraph{\emph{Structured notes}.}
Structured-note takeover provides a fixed-field handoff record before the
original issue. The experiment code computes the deterministic fields from
checkpoint metadata, event logs, and the frozen repository state. A
predecessor-side summarizer fills the remaining note fields from
predecessor-observable evidence. The model-filled fields are treated as
historical notes rather than ground truth.

\begin{quote}\small\raggedright
\texttt{=== Previous-agent structured handoff notes ===}\\
Structured handoff prepared from previous-agent evidence.\\
\textbf{Deterministic continuation-state fields:} repository change state;
changed source files; non-source artifacts observed; validation outcome after
latest source change; latest predecessor validation command; latest validation
evidence; continuation-state label.\\
\textbf{Model-generated note fields:} problem understanding; work completed;
evidence observed; observed failures or error evidence; remaining uncertainty;
rollback notes; recommended next action.\\
\texttt{=== End previous-agent structured handoff notes ===}\\
\texttt{=== Original task prompt ===}\\
\texttt{[ORIGINAL TASK PROMPT]}\\
\texttt{=== End original task prompt ===}
\end{quote}

Summary notes and the model-filled structured-note fields are generated from
predecessor-observable evidence: the task prompt, event log, command outputs,
and checkpoint repository state. They do not include official solutions, hidden
tests, or events after handoff.

\subsection{Example Structured Handoff}
\label{app:example-handoff}

The following excerpt illustrates the kind of information a structured handoff
passes to the successor. It is shortened for space, but preserves the distinction
between deterministic checkpoint state and model-generated predecessor notes.

\begin{quote}\small\raggedright
\texttt{=== Previous-agent structured handoff notes ===}\\
\textbf{Deterministic continuation state}\\
\texttt{Changed source files: [package/module.py]}\\
\texttt{Non-source artifacts: NONE OBSERVED}\\
\texttt{Latest validation command: pytest path/to/test.py -q}\\
\texttt{Latest validation evidence: failed assertion in edge-case behavior}\\
\texttt{Continuation state: unresolved; needs completion}\\[2pt]
\textbf{Previous-agent notes}\\
\texttt{Problem understanding: the issue concerns an edge case in input handling.}\\
\texttt{Work completed: adjusted the branch that normalizes the affected value.}\\
\texttt{Evidence observed: targeted validation still fails on the edge case.}\\
\texttt{Remaining uncertainty: whether the change preserves existing behavior.}\\
\texttt{Recommended next action: inspect the failing assertion, revise the source change, and rerun targeted validation.}\\
\texttt{=== End previous-agent structured handoff notes ===}
\end{quote}

Repository-only takeover receives none of this historical context; raw-trace
takeover receives the underlying event history instead of this bounded record.

\subsection{Observed Failure Modes}
\label{app:failure-modes}

Context-bearing handoffs are beneficial but not uniformly reliable. Manual inspection revealed three recurring failure modes. The first arises when a compact note preserves the high-level story while omitting an exact validation command or warning that matters for the next edit. The second occurs when a successor over-trusts a predecessor interpretation and continues along an unproductive path instead of rechecking the repository state. The third is specific to raw traces: they can contain enough low-level noise that the successor spends effort separating historical dead ends from useful evidence. These cases motivate the prompt instruction to treat handoff text as evidence to verify rather than as ground truth.

Matched cases from the Qwen-to-Qwen study illustrate these tradeoffs. In
\texttt{sphinx-doc\_\_sphinx-8265}, raw trace solves with 43 events while
repository-only, summary notes, and structured notes fail, showing a case where
exact trajectory history matters. In \texttt{sphinx-doc\_\_sphinx-8548},
structured notes solve while the other views fail; the note identifies the
inherited-attributes bug, the relevant test, and uncertainty in the predecessor's
partial fix, making the next verification step easier to follow. In
\texttt{pytest-dev\_\_pytest-10051}, raw trace and summary notes solve while
repository-only and structured notes fail; the structured record leaves
validation status unclear, showing the cost of a bounded record when it weakens
a crucial clue. Together these cases show that no single handoff view dominates:
structured notes help when validation evidence fits their fixed fields, and can
hurt when the bounded record drops a clue that free-form text or raw trace keeps.

\section{Use of AI Assistants}
We used AI assistants to help polish the writing and debug code. All ideas, analyses, claims, and final content presented in this paper remain the sole responsibility of the authors.
\end{document}